%% file: main.tex
\documentclass[journal,twoside,web]{IEEEtran}
\usepackage{cite}
\usepackage{amsmath,amssymb,amsfonts}
\usepackage{algorithmic}
\usepackage{graphicx}
\usepackage{url}
\usepackage{textcomp}
\usepackage{tikz}
\usepackage{orcidlink}
\usepackage{tabularx}
\usepackage{array}
\usepackage{xcolor}
\usetikzlibrary{arrows.meta, positioning, calc}

\begin{document}

\title{Region-Aware Multi-View Inspection Using Deep Learning and Domain-Informed Decision Fusion}

\author{
Srivishnu Ethirajulu Krishnaraj\,\orcidlink{0009-0005-8316-8128},
Rishikesh R. Chopade\,\orcidlink{0009-0008-8896-9487}
\thanks{Manuscript received October 21, 2026. This work was carried out as an internal project at Bosch Limited, India, which provided all resources. The system is deployed in series production. The dataset and codebase are proprietary to Bosch and are not publicly available.}
\thanks{S. E. Krishnaraj is with Bosch Limited, India (e-mail: Srivishnu.EthirajuluKrishnaraj@in.bosch.com, srivishnu.ek@gmail.com).}
\thanks{R. R. Chopade was an intern with Bosch Limited, India (e-mail: chopaderishikesh@gmail.com).}
}

\maketitle

\begin{abstract}
Automated optical inspection of precision components with tiny surface defects remains challenging when different regions of the same component exhibit different defect characteristics and, therefore, different acceptance criteria. This paper presents a region-aware, dual-model inspection framework that combines instance segmentation of localized seat-region defects with image-level classification of global shaft-region anomalies. Rather than treating these two model detections as sufficient evidence for component rejection, a deterministic decision fusion layer that formalizes industrial Standard Operating Procedures is applied to the model outputs to make the final component level decision. A PLC-synchronized imaging system acquires 18 seat and 18 shaft images over a full $360^{\circ}$ rotation to provide multi-view circumferential coverage. The framework was evaluated on 380 automotive fuel injector needles, covering representative operational defect types, using identical model predictions in all configurations. Direct model-output aggregation via simple union logic (where any single-view detection flags a rejection) identified all 104 non-conforming components but produced a component-level false rejection rate of 57.6\%. Inclusion of a complete decision layer maintained the observed non-conforming (NOK) recall of 100\% while reducing the false rejection rate to 10.9\%, corresponding to an 81.1\% reduction in false-positive component decisions. The results demonstrate that formalized manufacturing acceptance criteria can complement sensitive visual models by translating frame-level detections into practical and traceable component-level decisions.
\end{abstract}


\section{Introduction}
\label{sec:introduction}
\IEEEPARstart{H}{igh-precision} manufacturing of functionally critical components demands reliable quality inspection because small surface flaws may cause a component to violate its specified acceptance criteria. Inspection is particularly challenging for reflective or geometrically complex components because defect appearance can vary with viewpoint and imaging conditions~\cite{b15,b22}, while production deployment also requires reliable synchronization and timely decision transfer~\cite{b19}.

Deep learning has significantly improved industrial visual inspection by learning discriminative representations for subtle anomalies and enabling defect localization beyond image-level decisions~\cite{b3,b13}. However, practical deployment in high-throughput manufacturing still faces two recurring issues across many product types. First, highly sensitive detectors can trigger excessive false rejects when visually benign marks are flagged as defects, reducing line efficiency and operator trust~\cite{b10,b11}. Second, single-view inspection can miss defects that are only observable at certain angles on reflective or curved surfaces~\cite{b22}, motivating multi-view acquisition and fusion~\cite{b15}. In addition, serial two-stage architectures, in which a segmentation network first localizes potential defects and a subsequent decision network performs image-level defective/non-defective classification, have demonstrated strong surface-defect detection performance~\cite{b17}. However, such architectures differ from the present framework because both stages analyze the same inspected surface in sequence, whereas the proposed system assigns segmentation and classification to different functional regions and combines their outputs only at the component-decision level.

A key challenge in industrial visual inspection is that detecting a visual irregularity and determining component acceptability are not identical tasks. A sensitive model may identify small surface defects or benign visual patterns, but not every model detection necessarily violates the applicable component acceptance criteria. Its relevance may depend on its location relative to the component boundary, functional region, predicted category, dimensions, and occurrence count.

This distinction is particularly important when multiple functionally critical regions coexist within a single component. Such regions may exhibit different defect morphologies, functional risks, and acceptance conditions. Applying a uniform decision criterion across the entire component can therefore cause small, out-of-region, or functionally permissible detections to propagate into component-level false rejects. At the same time, some defects manifest as global structural anomalies that do not require pixel-level localization but must still be detected reliably. These considerations motivate a region-aware design in which specialized models generate visual detections and a separate decision layer evaluates those detections before component-level aggregation.

Rather than focusing solely on improving standalone visual models, manufacturers can complement model development by formalizing established SOPs and manufacturing acceptance standards as deterministic, traceable decision logic within the inspection system. In the proposed framework, the deep learning models generate frame-level defect detections, while the specification-derived decision layer determines whether those detections violate the applicable component acceptance criteria.

To implement this principle, this paper proposes a region-specific, dual-model inspection framework integrated into a PLC-synchronized multi-view imaging system. The approach employs (i) an instance segmentation model for fine-grained defects in a critical functional sub-region and (ii) a dedicated CNN classifier for global structural anomalies on the remaining component surfaces. A domain-informed decision-fusion layer evaluates the seat-region detections using specification-derived acceptance rules and aggregates the retained defects across views. The framework is evaluated using an automotive fuel injector needle as a representative industrial case study.

The main contributions of this paper are:
\begin{itemize}
    \item A \textbf{region-specific dual-model inspection pipeline} that allocates instance segmentation to localized defects in a critical functional sub-region and image-level classification to global anomalies on the remaining surfaces, aligning model choice with region-dependent defect semantics.
    \item A \textbf{PLC-synchronized multi-view acquisition setup} that inspects each component over a full $360^{\circ}$ rotation to improve viewpoint coverage of reflective or geometrically complex parts.
    \item A \textbf{domain-informed decision-fusion layer} that formalizes established manufacturing acceptance criteria through three functional rule groups and aggregates the retained defects across views to produce a traceable component-level decision.
    \item A \textbf{component-level industrial evaluation with functional rule-group ablation} demonstrating how the complete decision layer and its individual rule groups affect false rejects when the trained models and their frame-level predictions remain fixed.
\end{itemize}

\section{Related Work}
\label{sec:relatedwork}

This section reviews single-model inspection, hybrid pipelines, false-positive reduction, multi-view inspection, and vision--PLC integration before identifying the research gap addressed by the present work.

\subsection{Learning-Based Inspection with Single Models}
Early deep-learning inspection systems commonly rely on a single model for image-level classification or defect localization. CNN backbones such as ResNet~\cite{b14} and EfficientNetV2~\cite{b25} provide strong feature representations for industrial imagery, and have yielded robust results in practice. For example, El Hachem \textit{et al.}~\cite{b2} applied a ResNet-50 classifier to verify the presence, absence, or incorrect type of fasteners across multiple regions of an automotive subassembly, reporting an accuracy of 99\%. Transfer learning has also been applied: Sassi \textit{et al.}~\cite{b4} developed a DenseNet-based transfer-learning system for welding-defect inspection of fuel injectors. Their final selected model achieved 97.22\% accuracy and 100\% recall on the reported test set. For cases requiring spatial insight, region-based architectures enable precise localization of anomalies; Mask~R-CNN not only detects but also delineates defect regions~\cite{b3,b13}. Kumar \textit{et al.}~\cite{b5} applied Mask~R-CNN with a ResNet-101-FPN backbone to metal-surface crack segmentation. The authors reported 96.8\% accuracy and 88.9\% mAP, compared with 94.3\% accuracy and 84.7\% mAP for the evaluated YOLOv5 model. These works demonstrate the effectiveness of deep models for specific inspection tasks, but they often apply uniform decision criteria across an entire component. A standalone model may struggle when different regions demand different acceptance thresholds, or when defect manifestations range from tiny localized flaws to broad surface anomalies.

\subsection{Multi-Stage and Hybrid Defect Pipelines}
To handle diverse defect types and drive down error rates, many researchers have proposed multi-stage or hybrid inspection architectures. In such pipelines, an initial stage typically proposes candidate defects, which subsequent stages then verify, refine, or classify. For example, Schlosser \textit{et al.}~\cite{b6} combined traditional image processing for preliminary fault detection with a deep CNN classifier, achieving an F1-score above 99\% in semiconductor wafer inspection, substantially better than what either method achieved alone. Similarly, Chow \textit{et al.}~\cite{b7} implemented a three-step workflow comprising unsupervised anomaly detection, blob segmentation, and classification for concrete structure images, enhancing the robustness of crack detection while controlling false alarms. Yang \textit{et al.}~\cite{b8} combined morphological preprocessing, instance segmentation, and a subsequent classification model for X-ray inspection of flip-chip packaging. The preprocessing stage was used to suppress irrelevant structures and false-positive detections, and the authors reported an increase in overall detection accuracy from 95\% to 99\%. These examples show that dividing the inspection task into specialized sub-tasks can improve accuracy on complex surfaces. However, cascaded pipelines must be carefully designed to prevent error propagation, and most do not explicitly address region-specific acceptance criteria.

\subsection{False-Positive Reduction and Post-Processing}
Minimizing false alarms is essential for industrial deployment because excessive false rejects reduce production efficiency and operator confidence. Haddad \textit{et al.}~\cite{b10} proposed a three-stage semiconductor inspection framework comprising locally adaptive change detection, deep-learning-based false-positive rejection, and defect-type-dependent refinement. Their refined system achieved up to 97.88\% overall detection accuracy. The authors additionally reported 98.2\% precision and 99.44\% recall for the refinement-stage acceptance of non-defective units. Qiu \textit{et al.}~\cite{b11} introduced an object-level SVM post-processing classifier for filtering false-positive patches produced by unsupervised anomaly detectors. For Cflow on the TFDS-RP dataset, the reported image-level F1-score increased from 46.15\% to 78.26\%. These studies show that secondary verification and domain-dependent refinement can substantially reduce false alarms. However, such post-processing must be designed carefully because over-aggressive filtering may suppress genuine defects.

\subsection{Viewpoint Coverage: Multi-View and 360$^{\circ}$ Inspection}
Defects on reflective or curved components may be visible only from certain angles, motivating multi-view inspection strategies. Bhavanasi \textit{et al.}~\cite{b15} showed that early and late fusion of multiple viewpoints improved scratch-segmentation performance over a mono-view baseline. In parallel, 360$^{\circ}$ inspection pipelines have been proposed with multiple viewpoints to improve surface coverage; for example, Kuang \textit{et al.}~\cite{b16} reported an automated multi-camera system for unsupervised defect detection that operates in real production conditions. Other approaches integrate multi-view information at the feature level. Zhou \textit{et al.}~\cite{b22} introduced intra-view and cross-view attention to adaptively integrate features from multiple perspectives. Their method improved defect-detection and localization performance over the compared multiview approach on the Real-IAD dataset. While multi-view schemes improve coverage and robustness, they are often focused on learned feature fusion or only decision-level voting rather than functional region distinctions or region-specific acceptance criteria.

\subsection{Industrial Deployment and Vision--PLC Integration}
Production deployment requires reliable synchronization, deterministic communication, and traceable transfer of inspection decisions to the control system. Ma\'slanka \textit{et al.}~\cite{b19} integrated a smart camera with a PLC through Ethernet/IP and implemented triple verification of consecutive inspection outputs at the PLC level. The authors reported a 28\% reduction in false positives relative to camera-only classification. Vieira \textit{et al.}~\cite{b24} examined a different form of integration by executing conventional computer-vision algorithms directly on a PLCnext controller. Their results showed that PLC-based image processing was feasible for comparatively simple applications with undemanding cycle-time requirements, while computationally intensive operations remained substantially slower than on a conventional computer. These studies demonstrate the practical value of vision--PLC integration, but neither develops a functional region-specific acceptance policy of the type considered in the present work.

\subsection{Research Gap and Novelty}
Prior works have advanced deep-learning inspection along several directions, including \emph{multi-view fusion}~\cite{b15}, \emph{full 360$^{\circ}$ capture pipelines}~\cite{b16}, \emph{serial segmentation--classification architectures}~\cite{b17}, and \emph{vision--PLC integration}~\cite{b19}, but these are largely explored separately. The reviewed approaches do not jointly address \emph{(i) region-specific defect semantics}, \emph{(ii) complete multi-view detection aggregation}, and \emph{(iii) traceable, deterministic OK/NOK decision logic} in one inspection system. The present work addresses this gap by combining region-aware model allocation, multi-view aggregation, and specification-derived decision logic within one inspection system. The automotive fuel injector needle case study evaluates whether formalizing established acceptance criteria can improve component-level decisions when the trained models and their frame-level predictions remain unchanged.

\section{Proposed Methodology}
\label{sec:methodology}

\subsection{Overview and System Architecture}
\label{sec:overview}
The proposed inspection system is designed as a region-aware, dual-model pipeline integrated with multi-view acquisition and domain-informed decision fusion. The core design principle is that different regions of a precision component exhibit distinct defect semantics and functional criticalities, which cannot be reliably handled by a single uniform detection strategy.

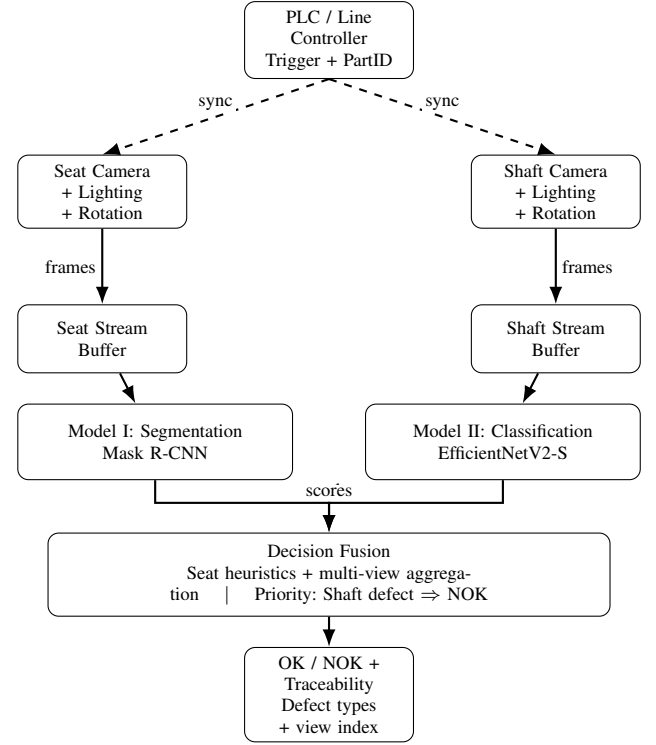
\begin{figure}[!htbp]
\centering
\input{system_arch.tikz}
\caption{Overall system architecture of the proposed PLC-synchronized, region-aware dual-model inspection pipeline (two camera views).}
\label{fig:system_arch}
\end{figure}

The system integrates (i) PLC-synchronized multi-view image acquisition over a full $360^{\circ}$ rotation, (ii) two specialized deep learning stages for complementary defect types, and (iii) a domain-informed decision-fusion layer that evaluates Model~I detections against specification-derived acceptance conditions before aggregating the retained defects across views. The overall architecture is illustrated in Fig.~\ref{fig:system_arch}.

\subsection{Region Definition and Inspection Scope}
\label{sec:regions}
In the considered automotive fuel-injector needle, the inspected surface is partitioned into two primary regions: the \textit{seat region} and the \textit{shaft region}. These regions exhibit different defect manifestations and functional risks and are therefore evaluated using separate inspection models and decision criteria.

\begin{figure}[!htbp]
\centering
\includegraphics[width=0.6\columnwidth]{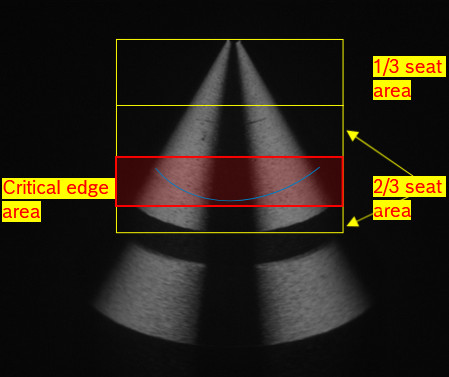}
\caption{Region-wise decomposition of the inspected component, including the upper one-third tip region, lower two-thirds seat region, and critical-line/edge region.}
\label{fig:regions}
\end{figure}

As shown in Fig.~\ref{fig:regions}, the seat region is further divided into an upper one-third region, corresponding to the tip of the component, and a lower two-thirds region containing the functionally critical sealing line. The lower two-thirds is further separated into regions above, around, and below the critical line to support location-dependent defect evaluation.

The seat-region decision layer evaluates each Model~I detection according to its position, functional region, predicted category, bounding-box dimensions, and occurrence count. Detections satisfying the applicable acceptance conditions are excluded from the component rejection decision, whereas detections exceeding those conditions are retained as defects.

To reflect their functional roles and support ablation analysis, Rules R1--R10 are organized into three groups.

\noindent\textbf{Group I: Size and Boundary Filtering (R1 and R4).}
This group excludes detections that do not satisfy the minimum size requirement or lie outside the applicable component boundary.
\begin{itemize}
    \item \textbf{R1 (Minimum detection-size criterion):} Detections with bounding-box dimensions not exceeding $8 \times 8$ pixels are excluded from the component rejection decision.
    \item \textbf{R4 (Component-boundary filtering):} Detections whose bounding boxes lie below the detected needle boundary are excluded from the component rejection decision.
\end{itemize}

\noindent\textbf{Group II: Region-Specific Defect Acceptance (R2, R3, and R5--R7).}
This group evaluates each detection according to its functional region, predicted category, dimensions, and the corresponding acceptance limit.
\begin{itemize}
    \item \textbf{R2 (Tip-region category filtering):} Within the upper one-third tip region, detections not classified as \textit{Tip Damage} are excluded from the component rejection decision.
    \item \textbf{R3 (Tip-damage retention):} Detections classified as \textit{Tip Damage} are retained and contribute to the seat-region decision.
    \item \textbf{R5 (Lower-region dimensional acceptance):} Within the lower two-thirds region (excluding critical-line zone), detections with dimensions not exceeding $28 \times 28$ pixels satisfy the applicable dimensional acceptance condition.
    \item \textbf{R6 (Grinding-mark dimensional acceptance):} Within the lower two-thirds region, detections classified as \textit{Grinding Marks} satisfy the applicable category-specific acceptance condition when their dimensions do not exceed $40 \times 30$ pixels.
    \item \textbf{R7 (Critical-line dimensional acceptance):} Within the critical-line region, a detection smaller than $18 \times 18$ pixels satisfies the applicable dimensional acceptance condition provided that no more than one such detection is present in the frame. (Multiplicity in this zone is governed in group III)
\end{itemize}

\noindent\textbf{Group III: Defect-Count Evaluation (R8--R10).}
This group determines whether detections satisfying individual dimensional conditions become rejectable because their combined count exceeds the applicable limit.
\begin{itemize}
    \item \textbf{R8 (Combined allowable-defect limit):}  Detections satisfying the dimensional acceptance conditions of R5, R6, or R7 are excluded from the component rejection decision when their combined count does not exceed five within the frame, subject to the zone restriction in R10.
    \item \textbf{R9 (Excessive allowable-defect count):} When the combined count of dimensionally accepted detections (under R5, R6, and R7) exceeds five within the frame, all such detections are retained and contribute to the component rejection decision.
    \item \textbf{R10 (Multiple critical-line defects):} When more than one detection satisfies the critical-line dimensional condition (R7), all critical-line detections are retained and contribute to the component rejection decision, regardless of whether the combined frame count in R8 is satisfied.
\end{itemize}

Rules R1--R10 were derived from pre-existing manufacturing quality standards and SOPs. This  enforces drawing tolerances unlike uncontrolled boundary learning of data-driven meta-classifiers. Requirements defined by physical dimensions, defect counts, categories, and functional locations were translated into deterministic machine-executable conditions.

To establish traceability between manufacturing drawing specifications in millimeters and the corresponding image-space rejection thresholds in pixels, in-situ target-based optical calibration was conducted using a precision metrology reticle. The calibration target was positioned at the component inspection datum plane under the fixed working distance and camera configuration used during seat-region inspection.

The pixel extents corresponding to known physical dimensions on the calibration target were measured to determine the effective optical scale in pixels per millimeter. This scale was subsequently used to convert the physical dimensional acceptance limits into the pixel-based conditions defined in Rules R1--R10. Consequently, these dimensional criteria represent calibrated image-space conversions of pre-existing manufacturing standards rather than post-hoc statistical parameters derived from the component-level evaluation dataset.

The shaft region is inspected separately using the Model~II classification model. Unlike the localized seat-region analysis, shaft inspection targets global rotational anomalies. A shaft frame classified as defective contributes to the final \textit{Rotation Bad} decision.

This rule-based evaluation prevents every model detection from being treated as equivalent component-level rejection evidence. Furthermore, unlike a black-box classifier, the rule-based method provides much-needed traceability in production. Detections satisfying the applicable size, boundary, regional, category, dimensional, and count conditions are excluded from the rejection decision. The remaining defects are passed to the multi-view aggregation stage described in Section~\ref{sec:fusion}.

\subsection{Multi-View Acquisition and Processing Flow}
\label{sec:multiview}
To provide circumferential coverage of the cylindrical and reflective component, the PLC-synchronized inspection station acquires images at 18 angular positions separated by $20^{\circ}$ over a complete $360^{\circ}$ rotation. At each position, the seat and shaft cameras acquire one image each, yielding 36 frames per component.

Due to corporate non-disclosure restrictions, absolute line-specific takt times are not reported. Relative to the pre-existing manual inspection baseline, the automated system reduced the per-component cycle time by 27.3\% and remained within the applicable internal manufacturing limit. Servo-based indexing and image acquisition accounted for approximately 55\% of the automated cycle, with the remainder comprising component handling and neural-network inference. The deterministic decision-fusion layer introduced negligible additional latency, indicating that its reduction of false rejects did not create a meaningful runtime bottleneck.

\begin{figure}[!htbp]
\centering
\includegraphics[width=0.5\linewidth]{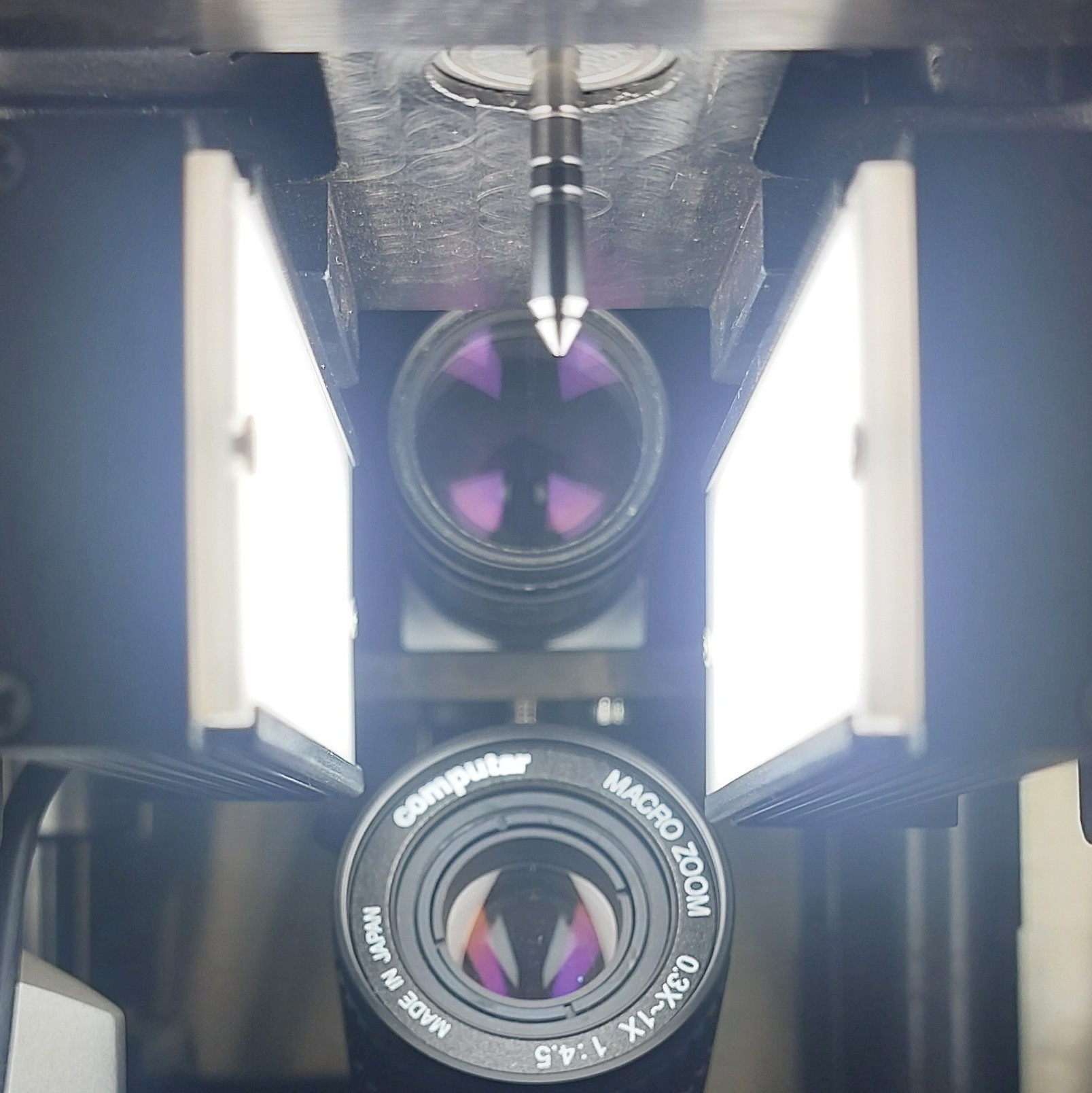}
\caption{Camera alignment and close-up view of the rotating component during inspection.}
\label{fig:alignment}
\end{figure}

Seat-region frames are processed by Model~I for localized defect segmentation, whereas shaft-region frames are processed by Model~II for global anomaly classification. The decision-fusion layer evaluates the seat-region detections and combines the retained defects with the shaft-region predictions across all available views. The acquisition configuration is summarized in Table~\ref{tab:acquisition_setup}, and Fig.~\ref{fig:alignment} shows the camera alignment and component positioning.

\begin{table}[!htbp]
\centering
\caption{Multi-view acquisition and inspection system configuration.}
\label{tab:acquisition_setup}
\small
\renewcommand{\arraystretch}{0.75}
\setlength{\tabcolsep}{5pt}
\begin{tabularx}{\columnwidth}{
>{\raggedright\arraybackslash}p{0.35\columnwidth}
>{\raggedright\arraybackslash}X}
\hline
\textbf{Parameter} & \textbf{Configuration} \\
\hline
Views per rotation & 18 views at 20$^{\circ}$ angular intervals \\
Seat camera & Basler ace U acA1300-200uc with Computar MLM-3XMP Lens \\
Shaft camera & Basler ace U acA1300-200uc with Fujifilm Fujinon HF35HA-1B lens \\
Lighting & $420 \times 20$~mm LED bar light \\
Rotary fixture & Servo-driven \\
Seat-region resolution & $1280 \times 1024$ pixels \\
Shaft-region resolution & $1280 \times 1024$ pixels \\
\hline
\end{tabularx}
\end{table}

\subsection{Model I: Seat Region Defect Segmentation}
\label{sec:model1}
Model~I performs instance segmentation of spatially localized defects in the seat region. The implementation uses Mask~R-CNN with a ResNet-50-FPN V2 backbone. For each input image, the model produces instance masks, bounding boxes, class labels, and confidence scores. These outputs are subsequently processed using the region-specific acceptance rules described in Section~\ref{sec:regions} before multi-view aggregation by the decision-fusion layer described in Section~\ref{sec:fusion}.

\subsubsection{Seat-Region Dataset}
\label{sec:seat_dataset}
The Model~I dataset comprised 2,621 annotated images covering eight seat-region defect categories, as summarized in Table~\ref{tab:seat_dataset}. The defect regions were manually annotated using the Label Studio brush interface. The annotations were exported in run-length encoded format and converted into single-channel indexed masks for model training. The connected regions within each class mask were treated as separate defect instances. Instances whose extracted width or height was less than four pixels were excluded when constructing the training targets as they fell below reliable annotation repeatability.

\begin{table}[!htbp]
\centering
\caption{Class-wise image distribution of the seat-region dataset.}
\label{tab:seat_dataset}
\normalsize
\renewcommand{\arraystretch}{1.0}
\setlength{\tabcolsep}{6pt}
\begin{tabularx}{\columnwidth}{
>{\raggedright\arraybackslash}X
>{\centering\arraybackslash}p{0.35\columnwidth}}
\hline
\textbf{Category} & \textbf{Number of Images} \\
\hline
Delamination & 1,099 \\
Grinding Marks & 172 \\
Droplet & 671 \\
Tip Damage & 222 \\
Grinding at Neck & 27 \\
Grinding Not Done & 74 \\
Loading Defect & 325 \\
Grinding Ring & 31 \\
\hline
\textbf{Total Defective Images} & \textbf{2,621} \\
\hline
\end{tabularx}
\end{table}

The class distribution reflects a practical constraint of industrial defect-data collection, as some defect categories occur less frequently during production, making it difficult to accumulate comparable amounts of high-quality examples across all classes. To preserve the calibrated optical scale required in the subsequent dimensional rule evaluation (Section~\ref{sec:regions}), scale-based augmentation was intentionally omitted. Furthermore, to ensure that subtle surface defect signatures are preserved as the part rotates, optical jitters were not applied. OK images were not used for training because surfaces other than those annotated as defects are actually OK, and the region proposal network and ROI alignment head naturally sample negative background proposals per frame, providing normal surface topography. Also, training extensively on defect-bearing frames maximizes gradient updates for scarce defect morphologies.

\subsubsection{Preprocessing and Data Partitioning}
\label{sec:seat_preprocessing}
Each image was converted to RGB format and transformed into a PyTorch tensor. Standard ImageNet normalization and interval scaling were managed by default Torchvision Mask R-CNN transformation pipeline, which automatically remaps the predicted bounding boxes and masks back to original pixel resolution for subsequent dimensional rule-based evaluation. The resulting dataset partition is presented in Table~\ref{tab:seat_split}.

The dataset was partitioned using class-stratified sampling into training, validation, and test subsets at an approximate 80:10:10 ratio (Table~\ref{tab:seat_split}), using a fixed random seed of 42 for reproducibility. Because each captured frame corresponds to a distinct physical component, this partition inherently prevents component-level data leakage across splits. To ensure unambiguous category representation during stratified splitting, each curated training frame was cataloged according to its single represented defect category, while the underlying Mask R-CNN architecture retained standard multi-class detection capabilities during inference. To prevent optimistic evaluation bias, the Model I dataset was strictly confined to offline model development. All end-to-end system evaluations were conducted on an independent, component-level dataset detailed in Section~\ref{sec:experiments}.

\begin{table}[!htbp]
\centering
\caption{Partition of the seat-region dataset.}
\label{tab:seat_split}
\normalsize
\renewcommand{\arraystretch}{1.0}
\setlength{\tabcolsep}{4pt}
\begin{tabularx}{\columnwidth}{
>{\raggedright\arraybackslash}X
>{\centering\arraybackslash}p{0.30\columnwidth}
>{\centering\arraybackslash}p{0.30\columnwidth}}
\hline
\textbf{Subset} & \textbf{Number of Images} & \textbf{Approximate Percentage} \\
\hline
Training & 2,099 & 80\% \\
Validation & 261 & 10\% \\
Test & 261 & 10\% \\
\hline
\textbf{Total} & \textbf{2,621} & \textbf{100\%} \\
\hline
\end{tabularx}
\end{table}

\subsubsection{Model Configuration and Training}
\label{sec:seat_training}
The Mask~R-CNN model was implemented using PyTorch and Torchvision. The network was initialized from a locally stored COCO-pretrained Mask~R-CNN ResNet-50-FPN V2 checkpoint. Its box-classification and mask-prediction heads were subsequently replaced with application-specific prediction heads. The weights were finetuned end-to-end due to image domain shift. The reported training and evaluation considered the eight defect categories represented in the Model~I dataset. The complete model and training configuration are summarized in Table~\ref{tab:model1_training}.

Model performance was evaluated after every training epoch using the validation subset. Rather than relying on threshold-agnostic COCO mean Average Precision (mAP), model selection was guided by the instance-level F1-score evaluated at a fixed confidence threshold of 0.50 and an IoU threshold of 0.50. This balances precision and recall for production, where fixed decision thresholds are required. For each epoch, true-positive, false-positive, and false-negative counts were accumulated across all validation images and defect categories to compute global precision and recall. The checkpoint yielding the highest validation F1-score was selected for testing and downstream component-level inference. Training was terminated when this validation F1-score failed to improve for ten consecutive epochs.

\begin{table}[!htbp]
\centering
\caption{Model and training configuration for Model~I.}
\label{tab:model1_training}
\small
\renewcommand{\arraystretch}{0.75}
\setlength{\tabcolsep}{5pt}
\begin{tabularx}{\columnwidth}{
>{\raggedright\arraybackslash}p{0.41\columnwidth}
>{\raggedright\arraybackslash}X}
\hline
\textbf{Parameter} & \textbf{Configuration} \\
\hline
Architecture & Mask~R-CNN ResNet-50-FPN V2 \\
Framework & PyTorch and Torchvision \\
Initialization & COCO-pretrained Mask~R-CNN V2 weights \\
Evaluated defect categories & Eight defect categories \\
Explicit preprocessing & None \\
Data augmentation & None \\
Split strategy & Part-disjoint class-stratified \\
Split ratio & 80:10:10 \\
Random seed & 42 \\
Optimizer & AdamW \\
Loss function & Multi-task loss \\
Initial learning rate & $1\times10^{-4}$ \\
Weight decay & $1\times10^{-4}$ \\
Batch size & 4 images \\
Maximum training length & 50 epochs \\
Learning-rate scheduler & ReduceLROnPlateau \\
Scheduler metric & Global instance-level validation F1-score \\
Scheduler patience & 3 epochs \\
Scheduler reduction factor & 0.5 \\
Minimum learning rate & $1\times10^{-7}$ \\
Early-stopping patience & 10 epochs \\
Checkpoint-selection & Highest global validation F1-score \\
Detection confidence threshold & 0.50 \\
Evaluation match threshold & Bounding-box IoU $\geq 0.50$ \\
Mask binarization threshold & 0.50 \\
GPU & NVIDIA RTX A2000, 12~GB \\
CPU & 12th Gen Intel Core i5-12600K \\
System RAM & 32~GB DDR5 \\
\hline
\end{tabularx}
\end{table}

During Model~I validation and testing, predictions with confidence scores below 0.50 were discarded, and the remaining predictions were processed in descending order of confidence. Matching was class-consistent and one-to-one. For each retained prediction, the unmatched ground-truth instance of the same class having the highest bounding-box IoU was selected as the candidate match. A prediction was counted as a true positive when the candidate IoU was at least 0.50. Each ground-truth instance could be matched to at most one prediction. Unmatched predictions were counted as false positives, and unmatched ground-truth instances were counted as false negatives.

For each matched true-positive instance, the predicted mask probability map was binarized using a threshold of 0.50. Mask IoU was then calculated between the resulting binary prediction and the corresponding ground-truth mask. Mean mask IoU was calculated across matched instances and was not used to determine whether a prediction constituted a true positive. After model development, the selected Model~I checkpoint was used to generate seat-region predictions for the separate component-level dataset described in Section~\ref{sec:experimental_dataset}. The held-out test performance of Model~I is reported and discussed in Section~\ref{sec:results_discussion}.

\subsection{Model II: Shaft Region Defect Classification}
\label{sec:model2}
Model~II detects global structural anomalies in the shaft region using image-level classification. In the considered case study, the primary defect category is \textit{Rotation Bad}, which indicates improper rotational alignment or bending of the component.

\subsubsection{Shaft-Region Dataset}
The shaft-region dataset comprised 7,933 images, each acquired from different components. Each image was assigned one of two labels: \textit{Rotation OK} or \textit{Rotation Bad}. The dataset contained 7,479 \textit{Rotation OK} images and 454 \textit{Rotation Bad} images. The resulting class distribution is presented in Table~\ref{tab:shaft_dataset}.

\begin{table}[!htbp]
\centering
\caption{Class distribution of the shaft-region dataset.}
\label{tab:shaft_dataset}
\normalsize
\renewcommand{\arraystretch}{1.0}
\setlength{\tabcolsep}{6pt}
\begin{tabularx}{\columnwidth}{
>{\raggedright\arraybackslash}X
>{\centering\arraybackslash}p{0.22\columnwidth}
>{\centering\arraybackslash}p{0.30\columnwidth}}
\hline
\textbf{Class} & \textbf{Images} & \textbf{Approximate Percentage} \\
\hline
Rotation OK & 7,479 & 94.3\% \\
Rotation Bad & 454 & 5.7\% \\
\hline
\textbf{Total} & \textbf{7,933} & \textbf{100\%} \\
\hline
\end{tabularx}
\end{table}

\subsubsection{Preprocessing and Data Partitioning}
The shaft images were converted to RGB format and resized to $224 \times 224$ pixels. Pixel intensities were normalized using the ImageNet mean and standard deviation.

Training-time augmentation consisted of brightness and contrast adjustments implemented using ColorJitter, with adjustment factors of 0.1 for both brightness and contrast. Geometric transformations were omitted to preserve the fixed mechanical axis and runout profile of the component. No augmentation was applied to the validation or test subsets.

The shaft-region dataset was partitioned into training, validation, and test subsets using an approximate 80:10:10 ratio. The partitioning was performed separately for each class to preserve the representation of both classes across all three subsets. A fixed random seed of 42 was used for reproducible split generation. The resulting dataset partition is summarized in Table~\ref{tab:shaft_split}.

\begin{table}[!htbp]
\centering
\caption{Partition of the shaft-region dataset.}
\label{tab:shaft_split}
\normalsize
\renewcommand{\arraystretch}{1.0}
\setlength{\tabcolsep}{4pt}
\begin{tabularx}{\columnwidth}{
>{\raggedright\arraybackslash}X
>{\centering\arraybackslash}p{0.21\columnwidth}
>{\centering\arraybackslash}p{0.24\columnwidth}
>{\centering\arraybackslash}p{0.15\columnwidth}}
\hline
\textbf{Subset} & \textbf{Rotation OK} & \textbf{Rotation Bad} & \textbf{Total} \\
\hline
Training & 5,983 & 363 & 6,346 \\
Validation & 747 & 45 & 792 \\
Test & 749 & 46 & 795 \\
\hline
\textbf{Total} & \textbf{7,479} & \textbf{454} & \textbf{7,933} \\
\hline
\end{tabularx}
\end{table}

\subsubsection{Model Configuration and Training}
Model~II used an EfficientNetV2-S convolutional neural network~\cite{b25} initialized with ImageNet-pretrained weights. Its original classification layer was replaced with an application-specific two-class output layer corresponding to the \textit{Rotation OK} and \textit{Rotation Bad} classes. Table~\ref{tab:model2_training} summarizes the Model~II and training configuration. Training and evaluation were performed using the same workstation and hardware configuration reported for Model~I in Table~\ref{tab:model1_training}.

Most of the pretrained network parameters were frozen during training to avoid overfitting. Only the final EfficientNetV2 feature block and the classification head were updated. The model was optimized using AdamW with differential learning rates. The classification head was assigned an initial learning rate of $1\times10^{-3}$, while the final feature block was assigned an initial learning rate of $1\times10^{-4}$.

Class-weighted cross-entropy loss (balanced inverse) was employed to account for the class imbalance between the \textit{Rotation OK} and \textit{Rotation Bad} classes. The class weights were calculated from the corresponding class frequencies in the training subset.

Model performance was evaluated on the validation subset after each training epoch. The checkpoint achieving the highest validation macro F1-score was selected for downstream inference. Early stopping was applied when the validation macro F1-score did not improve for seven consecutive epochs. A StepLR scheduler reduced the learning rates by a factor of 0.1 after every ten training epochs.

\begin{table}[!htbp]
\centering
\caption{Model and training configuration for Model~II.}
\label{tab:model2_training}
\small
\renewcommand{\arraystretch}{0.75}
\setlength{\tabcolsep}{5pt}
\begin{tabularx}{\columnwidth}{
>{\raggedright\arraybackslash}p{0.45\columnwidth}
>{\raggedright\arraybackslash}X}
\hline
\textbf{Parameter} & \textbf{Configuration} \\
\hline
Model & EfficientNetV2-S \\
Framework & PyTorch and Torchvision \\
Initialization & ImageNet-pretrained weights \\
Input resolution & $224 \times 224$ pixels \\
Input preprocessing & RGB conversion, resizing, and ImageNet normalization \\
Training augmentation & Brightness and contrast adjustment using ColorJitter \\
Classification task & Binary classification \\
Loss function & Class-weighted cross-entropy \\
Optimizer & AdamW \\
Initial classifier learning rate & $1\times10^{-3}$ \\
Initial feature-block learning rate & $1\times10^{-4}$ \\
Weight decay & $1\times10^{-4}$ \\
Batch size & 64 images \\
Maximum training length & 30 epochs \\
Learning-rate scheduler & StepLR \\
Scheduler step size & 10 epochs \\
Scheduler reduction factor & 0.1 \\
Split strategy & Part-disjoint class-stratified \\
Split ratio & 80:10:10 \\
Random seed & 42 \\
Model-selection criterion & Highest validation F1-score \\
Early-stopping patience & 7 epochs \\
\hline
\end{tabularx}
\end{table}

Unlike Model~I, Model~II does not require pixel-level localization because \textit{Rotation Bad} is treated as a global image-level anomaly. For each shaft image, the classifier produces a binary prediction corresponding to the \textit{Rotation OK} and \textit{Rotation Bad} classes. These frame-level predictions are subsequently aggregated across all captured angular views using the decision-fusion mechanism described in Section~\ref{sec:fusion}.

The held-out test performance of Model~II is reported and discussed in Section~\ref{sec:results_discussion}.

\subsection{Domain-Informed Decision Fusion}
\label{sec:fusion}
The final inspection decision is obtained through a domain-informed decision-fusion layer that converts the frame-level outputs of both models into a component-level OK/NOK decision. A Model~I detection does not automatically cause component rejection. Its location, predicted category, dimensions, and occurrence count are first evaluated against the applicable manufacturing acceptance conditions. The retained seat-region defects are then aggregated across views and combined with the shaft-region predictions.

\begin{figure}[!htbp]
\centering
\includegraphics[width=0.85\linewidth]{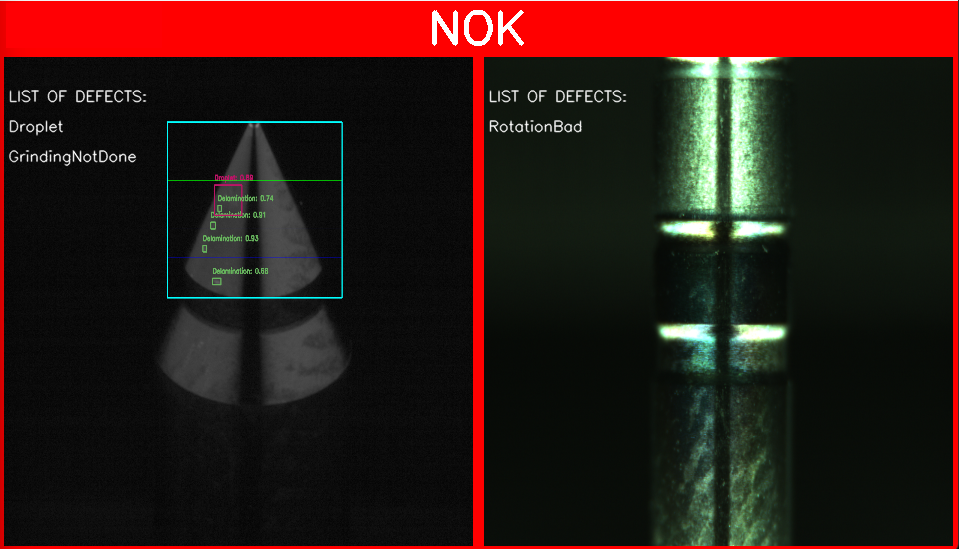}
\caption{Examples of seat- and shaft-region model outputs.}
\label{fig:seat_shaft_results}
\end{figure}

The decision layer is deterministic and does not employ learned fusion, probabilistic voting, or rule optimization using the component-level evaluation labels. Representative outputs of the two inspection models and the resulting component-level decision are shown in Fig.~\ref{fig:seat_shaft_results}.

\subsubsection{Seat-Region Rule Evaluation and Multi-View Aggregation}
Rules R1–R10 are evaluated on each seat-region frame in strict isolation, without requiring cross-view 3D defect tracking or multi-frame correspondence matching. On polished surfaces, viewpoint-dependent reflections render cross-view feature tracking fragile. Evaluating per frame level ensures deterministic, low-latency execution that replicates how human inspectors examine optical fields of view. Detections satisfying the applicable acceptance conditions are excluded from the component rejection decision, whereas the remaining defects are retained. The retained defect categories are then aggregated across all acquired seat-region views. If no retained defect remains after rule evaluation and aggregation, the seat region is classified as conforming. Otherwise, it is classified as non-conforming.

\subsubsection{Shaft-Region Fusion and Prioritization}
For the shaft region, the frame-level classifications are evaluated across all captured views. If at least three frames are classified as \textit{Rotation Bad}, the shaft region is classified as non-conforming. Otherwise, it is considered conforming. The shaft-region result takes precedence over the seat-region outcome because it is functionally critical. Furthermore, it acts as fail-fast logic in the production line.

\subsubsection{Final Part-Level Decision}
A non-conforming shaft-region outcome results in an immediate NOK decision. Otherwise, the seat-region outcome determines the final component status. The fusion layer also reports the retained defect categories and a representative frame index for visualization, logging, and traceability.

\section{Experiments}
\label{sec:experiments}

\subsection{Dataset and Inspection Setup}
\label{sec:experimental_dataset}
The experimental evaluation was conducted using an industrial dataset acquired from the PLC-synchronized inspection workbench described in Section~\ref{sec:multiview}. Each component was observed over a complete $360^{\circ}$ rotation, producing multiple images of the seat and shaft regions. Component-level ground-truth OK/NOK labels were obtained from the established manufacturing quality-assessment procedure independently of the predictions generated by the proposed inspection system.

The evaluation dataset contained 380 components, including 276 conforming components and 104 non-conforming components. The non-conforming set included components exhibiting localized seat-region defects, global shaft-region anomalies, or a combination of both. None of the components or images included in this final component-level evaluation dataset were used for training, validation, checkpoint selection, or held-out model-level testing of Model~I or Model~II. The component-level distribution is summarized in Table~\ref{tab:evaluation_dataset}.

\begin{table}[!htbp]
\centering
\caption{Component-level distribution of the evaluation dataset.}
\label{tab:evaluation_dataset}
\normalsize
\renewcommand{\arraystretch}{1.0}
\setlength{\tabcolsep}{6pt}
\begin{tabularx}{\columnwidth}{
>{\raggedright\arraybackslash}X
>{\centering\arraybackslash}p{0.35\columnwidth}}
\hline
\textbf{Component Status} & \textbf{Number of Components} \\
\hline
Conforming/OK & 276 \\
Non-conforming/NOK & 104 \\
\hline
\textbf{Total} & \textbf{380} \\
\hline
\end{tabularx}
\end{table}

\subsection{Compared Inspection Configurations}
\label{sec:compared_configurations}
As a part of Ablation study, five different component-level configurations were evaluated using identical Model~I and Model~II weights, inference settings, thresholds, and frame-level predictions. The \emph{Models Only} baseline directly aggregated the unfiltered model outputs using an any-defect rule. Three independent configurations applied only Group~I (R1 and R4), Group~II (R2, R3, and R5--R7), or Group~III (R8--R10) before component-level aggregation. Another configuration applied all retained rules R1--R10 in the \emph{Decision Layer}. 

The three group-wise configurations are independent rather than cumulative. For example, the Group~II configuration does not additionally apply Group~I. Their results therefore measure standalone group effects relative to the same baseline and should not be interpreted as additive.

\subsection{Implementation and Inference Protocol}
\label{sec:inference_protocol}
All configurations were evaluated using the same component-level dataset and identical frame-level predictions. The model weights, confidence thresholds, and inference settings were fixed before evaluation. No model retraining, rule modification, or configuration-specific threshold adjustment was performed using the component-level labels. Differences between configurations therefore arise only from the acceptance-rule groups applied before component-level aggregation.

\subsection{Evaluation Metrics}
\label{sec:evaluation_metrics}
Performance was evaluated at the component level, consistent with the requirements of industrial quality inspection. Let TP denote defective components correctly classified as NOK, FN denote defective components incorrectly classified as OK, FP denote conforming components incorrectly classified as NOK, and TN denote conforming components correctly classified as OK.

The component-level defect recall is defined as
\begin{equation}
\mathrm{Recall}_{\mathrm{NOK}}
=
\frac{\mathrm{TP}}
{\mathrm{TP}+\mathrm{FN}}.
\end{equation}

The false reject rate is defined as
\begin{equation}
\mathrm{FRR}
=
\frac{\mathrm{FP}}
{\mathrm{FP}+\mathrm{TN}}.
\end{equation}

Overall accuracy is calculated as
\begin{equation}
\mathrm{Accuracy}
=
\frac{\mathrm{TP}+\mathrm{TN}}
{\mathrm{TP}+\mathrm{TN}+\mathrm{FP}+\mathrm{FN}}.
\end{equation}

Because false rejects directly affect production throughput, the number of false rejects per 1,000 conforming components is additionally reported as
\begin{equation}
\mathrm{FR}_{1000}
=
1000 \times \mathrm{FRR}.
\end{equation}

NOK recall was treated as the primary defect-detection metric. Reductions in FRR were considered beneficial only when high component-level NOK recall was maintained.

\subsection{Results and Discussion}
\label{sec:results_discussion}

\subsubsection{Held-Out Model Performance and System-Level Interpretation} \label{sec:model_results} 
The held-out test results of the two visual models are summarized in Table~\ref{tab:model_test_results}. On the held-out test split of 261 seat-region images, Model~I achieved a precision of 53.24\% $(\mathrm{TP}=501,\ \mathrm{FP}=440)$, an instance-level recall of 37.19\% $(\mathrm{FN}=846)$, an F1-score of 43.79\%, and a mean bounding-box IoU of 0.7069 among correctly matched defect candidates. The observed instance-level recall of 37.19\% reflects the challenging imaging and dataset conditions associated with seat-region inspection. Both Model I and Model II were evaluated at the individual image level, not at the component level with all viewpoints and angles. On the polished metal surface, viewpoint-dependent specular reflections can affect defect contrast at glancing angles, causing an annotated defect to be missed in an individual image, while the same defect may be more clearly resolved from an adjacent angle where the illumination better reveals its surface topography. The class imbalance discussed in Section~\ref{sec:seat_dataset} may also have contributed to the missed detections, particularly for defect categories represented by fewer high-quality training examples.

\begin{table}[!htbp] 
\centering 
\caption{Held-out test performance of the two inspection models.} \label{tab:model_test_results} 
\small 
\renewcommand{\arraystretch}{0.95} 
\setlength{\tabcolsep}{4pt} 
\begin{tabularx}{\columnwidth}{ 
>{\raggedright\arraybackslash}p{0.18\columnwidth} 
>{\raggedright\arraybackslash}X 
>{\centering\arraybackslash}p{0.22\columnwidth}} 
\hline 
\textbf{Model} & \textbf{Metric} & \textbf{Result} \\ 
\hline 
Model~I & Instance-level F1-score & 43.79\% \\ 
Model~I & Instance-level recall & 37.19\% \\ 
Model~I & Mean bounding-box IoU$^{a}$ & 0.7069 \\ 
Model~II & Image-level accuracy & 99.62\% \\ 
Model~II & Macro F1-score & 98.32\% \\ 
Model~II & Rotation Bad recall & 100\% $(46/46)$ \\ 
\hline 

\multicolumn{3}{l}{\footnotesize $^{a}$Calculated over correctly matched Model~I detections.} 
\end{tabularx} 
\end{table} 

This viewpoint-dependent behavior motivates the use of Model~I as a localized candidate-generation stage within the multi-view inspection system. Each component is inspected at 18 angular positions separated by $20^{\circ}$, and the detected candidates are aggregated before the component-level decision is produced. The instance-level evaluation penalizes every unmatched annotated defect instance in an individual image, whereas the component-level evaluation determines whether sufficient rejectable evidence is available from the captured views to identify the complete component as non-conforming. Consequently, exhaustive detection of every annotated instance is not required for a NOK component to be correctly identified when relevant defect evidence is retained from at least one available view.

Importantly, for correctly matched defect candidates $(\mathrm{IoU} \geq 0.50)$, Model~I achieved a mean bounding-box IoU of 0.7069. In the context of the minute industrial surface anomalies considered in this study, whose limited spatial extent and visually diffuse boundaries under specular reflection make precise localization challenging, this value indicates close spatial agreement between the predicted and reference bounding boxes. The localization result is particularly relevant to the proposed decision-fusion layer because Rules R1--R10 evaluate the predicted bounding-box position and dimensions against calibrated pixel-based acceptance thresholds derived from the applicable manufacturing specifications. These geometric outputs support minimum-size filtering, component-boundary filtering, functional-region assignment, dimensional evaluation, and defect-count analysis.

The suitability of Model~I for the proposed framework was therefore assessed from its localization performance and its intended role as a candidate-generation stage, rather than by treating its standalone instance-level recall as satisfactory. Its ability to provide localized defect candidates from multiple angular views enables the subsequent decision layer to evaluate the position, dimensions, category, and occurrence count of the detected defects. The performance of the complete framework at the component level is evaluated separately in Section~\ref{sec:component_results}.

In contrast, Model~II demonstrated strong held-out image-level performance, achieving an accuracy of 99.62\% and a macro F1-score of 98.32\%. The model correctly classified all 46 \textit{Rotation Bad} images in the test subset. The minority-class support is reported because the perfect test recall was observed over a limited number of \textit{Rotation Bad} examples.

\subsubsection{Component-Level Performance and Rule-Group Ablation} \label{sec:component_results} 

\begin{table}[!htbp] 
\centering 
\caption{Component-level functional rule-group ablation results.} 
\label{tab:group_ablation} 
\scriptsize 
\renewcommand{\arraystretch}{1.08} 
\setlength{\tabcolsep}{2.2pt} 
\begin{tabularx}{\columnwidth}{ 
>{\raggedright\arraybackslash}X c c c c c} 
\hline 
\textbf{Metric} & 
\textbf{MO} & 
\textbf{G-I} & 
\textbf{G-II} & 
\textbf{G-III} & 
\textbf{CDL} \\ 
\hline 
TP & 104 & 104 & 104 & 104 & 104 \\ 
FN & 0 & 0 & 0 & 0 & 0 \\
FP & 159 & 120 & 40 & 114 & 30 \\ 
TN & 117 & 156 & 236 & 162 & 246 \\
NOK recall (\%) & 100.0 & 100.0 & 100.0 & 100.0 & 100.0 \\ 
FRR (\%) & 57.6 & 43.5 & 14.5 & 41.3 & 10.9 \\ 
Accuracy (\%) & 58.2 & 68.4 & 89.5 & 70.0 & 92.1 \\ 
FR/1000 OK & 576 & 435 & 145 & 413 & 109 \\ 
\hline 
\end{tabularx} 
\vspace{1mm} 
\begin{minipage}{\columnwidth} 
\footnotesize 
MO: Models Only; G-I: Group~I only; G-II: Group~II only; G-III: Group~III only; CDL: Complete Decision Layer (R1--R10). 
\end{minipage} 
\end{table}

Table~\ref{tab:group_ablation} reports the functional rule-group ablation results for the component-level evaluation. Models Only identified all 104 non-conforming components, corresponding to an observed NOK recall of 100\%. However, direct aggregation of the unfiltered model outputs incorrectly rejected 159 of the 276 conforming components, resulting in an FRR of 57.6\%. Thus, although the fixed model outputs provided sufficient evidence to identify every NOK component in the evaluation dataset, direct any-defect aggregation produced a high number of false rejects.

Applied independently, Group~I reduced FP from 159 to 120 and FRR from 57.6\% to 43.5\%. Group~III reduced FP to 114 and FRR to 41.3\%. These results show that size and boundary filtering and defect-count evaluation each provided measurable standalone reductions in false rejects. 

Group~II produced the largest standalone improvement, reducing FP from 159 to 40 and FRR from 57.6\% to 14.5\%. This result indicates that evaluating detections according to their functional region, predicted category, and applicable dimensional limits accounted for most of the standalone false-reject reduction. 

The complete R1--R10 decision layer achieved the strongest overall result, reducing FP to 30 and FRR to 10.9\%. This corresponds to 129 fewer false rejects and an 81.1\% relative reduction compared with Models Only. The complete decision layer also produced ten fewer false rejects than Group~II alone. Because the independently evaluated rule groups can affect overlapping component detections, their standalone effects are not additive. 

TP and FN remained unchanged at 104 and 0 across all configurations, corresponding to an observed NOK recall of 100\%. Accuracy increased from 58.2\% for Models Only to 92.1\% for the Complete Decision Layer, while the expected number of false rejects per 1,000 conforming components decreased from 576 to 109. Because all configurations used identical model weights, inference settings, thresholds, and frame-level predictions, these improvements resulted solely from how the fixed model detections were interpreted before component-level aggregation. 

These results demonstrate that formalizing established SOPs and manufacturing acceptance criteria can substantially reduce false rejects without removing the evidence required to identify the NOK components represented in the evaluation dataset. While the observed 100\% recall demonstrated strong detection robustness across evaluated components, sustained performance relies on maintaining stable optical conditions and adapting inspection pipeline to emerging defect morphologies. 

\section{Conclusion}
This paper presented a region-aware, dual-model framework for multi-view inspection of precision industrial components. The framework combines Mask R-CNN instance segmentation of localized seat-region defects, EfficientNetV2-S image-level classification of shaft-region anomalies, PLC-synchronized $360^{\circ}$ image acquisition, and deterministic decision fusion. The decision layer formalizes established manufacturing acceptance criteria through three functional rule groups covering size and boundary filtering, region-specific defect acceptance, and defect-count evaluation. 

On an independent industrial dataset of 380 components, direct aggregation of raw model outputs yielded an unacceptable 57.6\% false reject rate. Incorporating the complete decision layer reduced the false reject rate to 10.9\% while maintaining 100\% defect recall across all 104 non-conforming parts. Crucially, because all the evaluations utilized identical model predictions, gains stemmed entirely from specification-driven decision logic.

These findings show that industrial inspection development need not focus exclusively on further improving standalone model performance. When established SOPs and acceptance standards define the functional relevance of detected defects, these requirements can be formalized as deterministic and traceable decision logic within the inspection system. Such logic does not replace the need for reliable models, but complements them by connecting frame-level detections with practical component-level acceptance decisions. Future work will investigate adaptive mechanisms in decision-fusion heuristics to accommodate emerging defect morphologies in evolving production lines. 


\end{document}

%% file: system_arch.tikz
\centering
\begin{tikzpicture}[
  font=\scriptsize,
  box/.style={draw, rounded corners, align=center, minimum height=9mm, text width=20mm},
  sbox/.style={draw, rounded corners, align=center, minimum height=10mm, text width=34mm},
  fuse/.style={draw, rounded corners, align=center, minimum height=11mm, text width=72mm},
  arrow/.style={-Latex, thick},
  darrow/.style={-Latex, thick, dashed},
  lab/.style={fill=white, inner sep=1pt, font=\scriptsize, align=center}
]

\node[box] (plc) {PLC / Line Controller\\Trigger + PartID};

\node[box, below=10mm of plc, xshift=-30mm] (seatcam) {Seat Camera\\+ Lighting \\+ Rotation};
\node[box, below=10mm of plc, xshift=+30mm] (shaftcam) {Shaft Camera\\+ Lighting \\+ Rotation};

\node[box, below=10mm of seatcam] (seatbuf) {Seat Stream\\Buffer};
\node[box, below=10mm of shaftcam] (shaftbuf) {Shaft Stream\\Buffer};

\node[sbox, below=4mm of seatbuf, xshift=7mm] (Model1) {Model I: Segmentation\\Mask R-CNN};
\node[sbox, below=4mm of shaftbuf, xshift=-7mm] (Model2) {Model II: Classification\\EfficientNetV2-S};

\node[fuse, below=12mm of $(Model1)!0.5!(Model2)$] (fusion)
{Decision Fusion\\
\strut Seat heuristics + multi-view aggregation \quad $\mid$ \quad Priority: Shaft defect $\Rightarrow$ NOK};

\node[box, below=4mm of fusion] (out) {OK / NOK + Traceability\\Defect types + view index};

\draw[darrow] (plc.south) -- node[lab, left=2pt, above=1pt]{sync} (seatcam.north);
\draw[darrow] (plc.south) -- node[lab, right=2pt, above=1pt]{sync} (shaftcam.north);

\draw[arrow] (seatcam) -- node[lab, left=1pt]{frames} (seatbuf);
\draw[arrow] (shaftcam) -- node[lab, right=1pt]{frames} (shaftbuf);

\draw[arrow] (seatbuf) -- (Model1);
\draw[arrow] (shaftbuf) -- (Model2);

\draw[arrow] (Model1.south) -- ++(0,-3mm) -| node[lab, above=1pt]{masks} (fusion.north);
\draw[arrow] (Model2.south) -- ++(0,-3mm) -| node[lab, above=1pt]{scores} (fusion.north);

\draw[arrow] (fusion) -- (out);

\end{tikzpicture}

%% file: main.bbl
\begin{thebibliography}{00}
\bibitem{b15} G. Bhavanasi, D. Neven, M. Arteaga, S. Ditzel, S. Dehaeck, and A. Bey-Temsamani, ``Enhanced Vision-Based Quality Inspection: A Multiview Artificial Intelligence Framework for Defect Detection,'' \textit{Sensors}, vol. 25, no. 6, Art. no. 1703, Mar. 2025, doi: 10.3390/s25061703.

\bibitem{b22} J. Zhou, M. Liu, Y. Ma, S. Jiang, and Y. Wang, ``Multi-View Attention Guided Feature Learning for Unsupervised Surface Defect Detection,'' \textit{IEEE/ASME Trans. Mechatron.}, vol. 30, no. 4, pp. 2844--2852, Aug. 2025, doi: 10.1109/TMECH.2025.3566311.

\bibitem{b19} M. Ma\'slanka, D. Jancarczyk, and J. Rysi\'nski, ``Integration of Machine Vision and PLC-Based Control for Scalable Quality Inspection in Industry 4.0,'' \textit{Sensors}, vol. 25, no. 20, Art. no. 6383, Oct. 2025, doi: 10.3390/s25206383.

\bibitem{b3} M. K. Ferguson, R. Ak, Y.-T. T. Lee, and K. H. Law, ``Detection and Segmentation of Manufacturing Defects with Convolutional Neural Networks and Transfer Learning,'' \textit{Smart Sustain. Manuf. Syst.}, vol. 2, no. 1, pp. 137--164, Oct. 2018, doi: 10.1520/SSMS20180033.

\bibitem{b13} K. He, G. Gkioxari, P. Doll\'{a}r, and R. Girshick, ``Mask R-CNN,'' in \textit{Proc. IEEE Int. Conf. Comput. Vis. (ICCV)}, Venice, Italy, 2017, pp. 2961--2969, doi: 10.1109/ICCV.2017.322.

\bibitem{b10} B. M. Haddad, S. F. Dodge, L. J. Karam, N. S. Patel, and M. W. Braun, ``Locally Adaptive Statistical Background Modeling With Deep Learning-Based False Positive Rejection for Defect Detection in Semiconductor Units,'' \textit{IEEE Trans. Semicond. Manuf.}, vol. 33, no. 3, pp. 357--372, Aug. 2020, doi: 10.1109/TSM.2020.2998441.

\bibitem{b11} J. Qiu, H. Shi, Y. Hu, and Z. Yu, ``Enhancing Anomaly Detection Models for Industrial Applications through SVM-Based False Positive Classification,'' \textit{Appl. Sci.}, vol. 13, no. 23, Art. no. 12655, Nov. 2023, doi: 10.3390/app132312655.

\bibitem{b17} J. Bo\v{z}i\v{c}, D. Tabernik, and D. Sko\v{c}aj, ``End-to-End Training of a Two-Stage Neural Network for Defect Detection,'' in \textit{Proc. 25th Int. Conf. Pattern Recognit. (ICPR)}, Milan, Italy, Jan. 2021, pp. 5619--5626, doi: 10.1109/ICPR48806.2021.9412092.

\bibitem{b14} K. He, X. Zhang, S. Ren, and J. Sun, ``Deep Residual Learning for Image Recognition,'' in \textit{Proc. IEEE Conf. Comput. Vis. Pattern Recognit. (CVPR)}, Las Vegas, NV, USA, Jun. 2016, pp. 770--778, doi: 10.1109/CVPR.2016.90.

            
\bibitem{b25} M. Tan and Q. V. Le, ``EfficientNetV2: Smaller Models and Faster Training,'' in \textit{Proc. 38th Int. Conf. Mach. Learn. (ICML)}, virtual, Jul. 2021, vol. 139, pp. 10096--10106.



\bibitem{b2} C. El Hachem, G. Perrot, L. Painvin, and R. Couturier, ``Automation of Quality Control in the Automotive Industry Using Deep Learning Algorithms,'' in \textit{Proc. Int. Conf. Comput., Control Robot. (ICCCR)}, Shanghai, China, Jan. 2021, pp. 123--127, doi: 10.1109/ICCCR49711.2021.9349273.


\bibitem{b4} P. Sassi, P. Tripicchio, and C. A. Avizzano, ``A Smart Monitoring System for Automatic Welding Defect Detection,'' \textit{IEEE Trans. Ind. Electron.}, vol. 66, no. 12, pp. 9641--9650, Dec. 2019, doi: 10.1109/TIE.2019.2896165.
            

\bibitem{b5} B. R. P. Kumar, B. S. Goud, N. Sangameshwar, D. Badrigari, K. Swaraja, and V. Himabindu, ``Automated Crack Defect Detection on Metal Surfaces in Manufacturing,'' in \textit{Proc. 6th Int. Conf. Inventive Res. Comput. Appl. (ICIRCA)}, Coimbatore, India, Jun. 2025, pp. 22--27, doi: 10.1109/ICIRCA65293.2025.11089523.
            

\bibitem{b6} T. Schlosser, M. Friedrich, F. Beuth, and D. Kowerko, ``Improving Automated Visual Fault Inspection for Semiconductor Manufacturing Using a Hybrid Multistage System of Deep Neural Networks,'' \textit{J. Intell. Manuf.}, vol. 33, no. 4, pp. 1099--1123, Apr. 2022, doi: 10.1007/s10845-021-01906-9.

\bibitem{b7}  J. K. Chow, Z. Su, J. Wu, P. S. Tan, X. Mao, and Y.-H. Wang, ``Artificial Intelligence-Empowered Pipeline for Image-Based Inspection of Concrete Structures,'' \textit{Autom. Constr.}, vol. 120, Art. no. 103372, Dec. 2020, doi: 10.1016/j.autcon.2020.103372.

\bibitem{b8} C.-H. Yang \emph{et al.}, ``High-Precision X-ray Inspection of Flip Chip Packaging Leveraging Deep Learning and Instance Segmentation for Manufacturing Process Optimization,'' in \textit{Proc. 26th Int. Conf. Electron. Packag. Technol. (ICEPT)}, Shanghai, China, Aug. 2025, doi: 10.1109/ICEPT67137.2025.11157128.

\bibitem{b16} Z. Kuang, X. Tie, L. Ying, and S. Jin, ``Computer Vision and Normalizing Flow-Based Defect Detection,'' \textit{arXiv preprint arXiv:2012.06737}, ver. 4, Feb. 2022, doi: 10.48550/arXiv.2012.06737.

\bibitem{b24} R. Vieira, D. Silva, E. Ribeiro, L. Perdigoto, and P. J. Coelho, ``Performance Evaluation of Computer Vision Algorithms in a Programmable Logic Controller: 
An Industrial Case Study,'' \textit{Sensors}, vol. 24, no. 3, Art. no. 843, Jan. 2024, doi: 10.3390/s24030843.
\end{thebibliography}
